\documentclass{Interspeech}

\interspeechcameraready

\title{Leveraging Unlabeled Audio-Visual Data in Speech Emotion Recognition using Knowledge Distillation}

\author[affiliation={}]{Varsha}{Pendyala}
\author[affiliation={}]{Pedro}{Morgado}
\author[affiliation={}]{William}{Sethares}

\affiliation{Department of Electrical and Computer Engineering}{University of Wisconsin-Madison}{USA}
\email{pendyala@wisc.edu, pmorgado@wisc.edu, sethares@wisc.edu}

\newcommand{\empfont}[1]{{\fontfamily{lmtt}\selectfont #1}}
\usepackage{comment}
\usepackage{multirow}
\begin{document}

\maketitle
\begin{abstract}
Voice interfaces integral to the human-computer interaction systems can benefit from speech emotion recognition (SER) to customize responses based on user emotions. Since humans convey emotions through multi-modal audio-visual cues, developing SER systems using both the modalities is beneficial. However, collecting a vast amount of labeled data for their development is expensive. This paper proposes a knowledge distillation framework called LightweightSER (LiSER) that leverages unlabeled audio-visual data for SER, using large teacher models built on advanced speech and face representation models. LiSER transfers knowledge regarding speech emotions and facial expressions from the teacher models to lightweight student models. Experiments conducted on two benchmark datasets, RAVDESS and CREMA-D, demonstrate that LiSER can reduce the dependence on extensive labeled datasets for SER tasks.

    
    
\end{abstract}

\section{Introduction}

Human-computer interaction systems equipped with voice interfaces are increasing in popularity. Detecting emotional states through spoken language, called speech emotion recognition (SER), is critical to effectively implement these systems. However, accurate SER is challenging due to the differences in accents, age, gender, and voice characteristics of the users. Human facial expressions and body language are closely linked to emotional states. Recent research~\cite{goncalves2022improving,Shukla_2023,Hajavi_2024} has shown that these visual cues can be used to enhance the accuracy of SER systems. However, collecting large volumes of manually labeled emotion data to develop accurate SER systems is both costly and time-consuming, largely due to the inherent ambiguity in humans' perception of emotions.

Recently, there has been significant progress in the field of audio, vision, and text, particularly in developing self-supervised learning (SSL) models such as HuBERT~\cite{Hsu_2021}, VideoMAE~\cite{Tong_2024}, and BERT~\cite{devlin-etal-2019-bert}. These models can be pre-trained on vast amounts of unlabeled data and subsequently fine-tuned using a limited quantity of task-specific labeled data, to yield remarkable performance in applications like facial expression recognition (FER)~\cite{sun2023mae} and SER~\cite{goron2024improving}. However, their large size makes these SSL models challenging to deploy in low-resource environments, such as mobile devices with computing and memory constraints. To overcome these challenges, knowledge distillation techniques~\cite{Hinton2015distill} are used to transfer the knowledge from large and accurate ``teacher'' models to lightweight ``student'' models. 
In these techniques, the student models are trained by aligning their intermediate feature representations or softmax distributions with those of the teacher.

Various distillation techniques have been explored in SER research, utilizing teacher models from the speech modality and other modalities such as vision and text. In~\cite{lou2024cubic, liu23b_interspeech}, the authors developed distillation techniques for speech SSL models that have been fine-tuned for SER. The authors in~\cite{shome2024speech} utilized cross-modal distillation from prosodic and linguistic teachers to boost the accuracy of their SER model. Another approach in~\cite{li2021speech} trained a student model on unlabeled audio-text pairs through cross-modal distillation from a strong BERT-based teacher that was fine-tuned on a text emotion corpus. In~\cite{Hajavi_2024}, SER models were developed using ground-truth labels and distillation from video models trained from scratch on labeled audio-visual data. However, no reported literature investigates distillation using unlabeled audio-visual data for SER.

The use of unlabeled audio-visual data to boost the performance of SER models has been reported in~\cite{goncalves2022improving,Shukla_2023}. In~\cite{goncalves2022improving}, the authors introduced an SSL framework, proposing new audio-visual pretext tasks to enhance speech representations for SER tasks. These cross-modal pretext tasks involve using acoustic features to predict the temporal variance of facial landmark positions, and multi-class pseudo-emotional labels derived from a combination of facial action units (AUs). However, relying solely on landmark variance prediction tasks or employing hand-engineered rules for generating pseudo-labels from AUs may not adequately capture the intricate changes in facial expressions over time. The authors in~\cite{Shukla_2023} train SER models through visual self-supervision via a face reconstruction task. In that approach, a speech encoder is jointly trained with a face encoder-decoder network to reconstruct video from a still face image paired with the corresponding speech utterance. However, the compute-intensive nature of this task presents significant challenges when attempting to scale this framework to large volumes of audio-visual data from everyday interactions.

This paper introduces LiSER, a knowledge distillation framework that utilizes unlabeled audio-visual data alongside a limited amount of labeled speech emotion data to build lightweight SER models. Our framework integrates state-of-the-art speech and face representation models to enhance the performance of lightweight SER models. We leverage unlabeled audio-visual data through the distillation of speech emotion knowledge from the HuBERT model, which has been fine-tuned for the SER task, while also incorporating insights from S2D~\cite{chen2024static}, a dynamic facial expression recognition (DFER) model. The DFER model is capable of recognizing facial expressions from raw pixel data in dynamic face image sequences or videos. As a result, our approach can efficiently leverage large-scale audio-visual data available on video-sharing platforms, employing the standard preprocessing pipeline typically associated with face recognition systems~\cite{du2022elements}.

We use the MSP-Face corpus~\cite{Vidal_2020} containing audio-visual data to extract emotion-related knowledge from HuBERT and S2D models. We train a lightweight SER model by employing both uni-modal and cross-modal distillation. In addition, we propose a novel training objective that incorporates instance-level confidence pertaining to emotion predictions of the teacher models. Systematic evaluations conducted on the RAVDESS~\cite{livingstone2018ryerson} and CREMA-D~\cite{Cao_2014} benchmarks yield several key findings: 1) Distillation from both audio and visual modalities of unlabeled data enhances the accuracy of the lightweight SER model 2) Utilizing both audio and visual modalities during the distillation process provides greater performance improvements compared to relying solely on one modality. 3) The integration of instance-level confidence related to the emotion predictions of teacher models shows promise for further enhancing the SER accuracy.

\section{Method}
\begin{figure}[!b]
  \centering
  \includegraphics[width=0.4\textwidth]{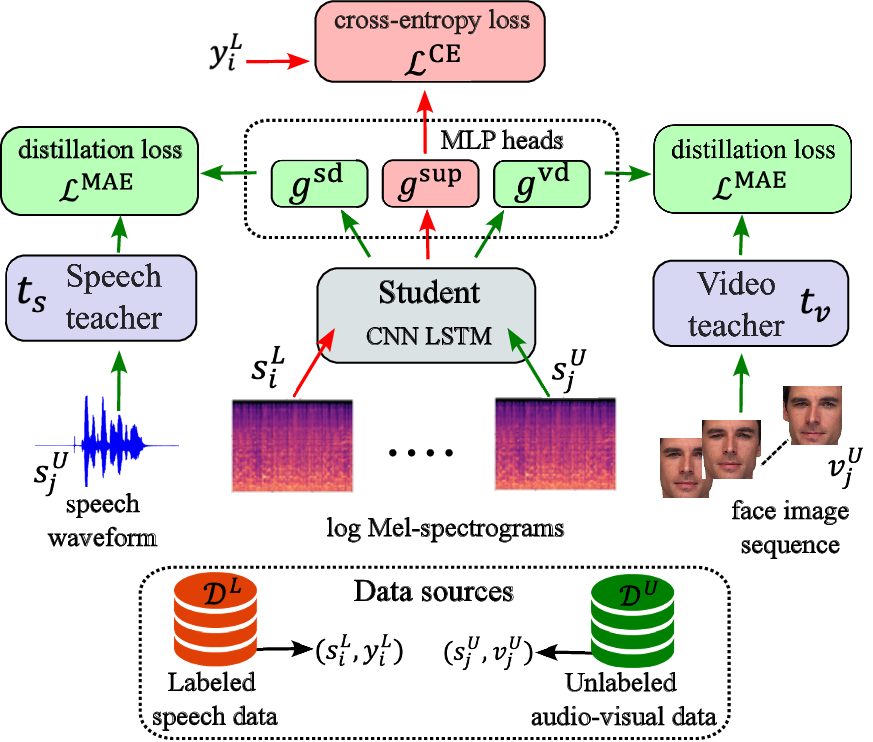}
  \caption{In LiSER, the student is trained using labeled speech through cross-entropy loss (red arrowed path), and unlabeled audio-visual data through distillation (green arrowed path).}
  \label{fig:emo_distill}
\end{figure}

This section outlines our approach, called LiSER, for training a lightweight SER model by leveraging unlabeled audio-visual data and a limited amount of labeled speech emotion data. Figure~\ref{fig:emo_distill} depicts the overall framework. 

\subsection{Speech teacher}
To develop a teacher model capable of identifying emotions from speech with high accuracy, LiSER starts with a HuBERT model~\cite{Hsu_2021} trained using self-supervised learning on a large corpus of unlabeled speech data. 
HuBERT's representations have proven to be beneficial across various applications, including speech recognition~\cite{Hsu_2021,Wang_2023}, speaker verification~\cite{Chen_2022}, and emotion recognition~\cite{Morais_2022,goron2024improving}. The model utilizes a convolutional encoder to capture local temporal features from raw speech inputs, along with a transformer encoder that generates global contextualized representations. We selected the base variant of the pre-trained HuBERT (hubert-base-ls960) from the HuggingFace library~\cite{wolf-etal-2020-transformers} and fine-tuned it for SER using the available labeled speech emotion data. The resulting speech teacher model processes raw speech waveforms as inputs and outputs the softmax probabilities corresponding to the emotion categories in the labeled speech.

\subsection{Video teacher}
LiSER uses a state-of-the-art dynamic facial expression recognition model (DFER) known as S2D~\cite{chen2024static} as a second teacher to extract emotional knowledge from videos. The architecture of S2D is based on the Vision Transformer (ViT) ~\cite{dosovitskiy2021an}. In~\cite{chen2024static}, the authors pre-train a ViT model to recognize facial expressions from static images, utilizing features derived from MobileFaceNet~\cite{Chen_2018}, a network designed for facial landmark detection. Subsequently, they adapt the static FER model for the dynamic FER task by training spatio-temporal adapters with face image sequence data obtained from the DFER dataset. S2D processes 16-frame face image sequences as inputs and produces softmax probabilities across the emotion categories found in the DFER dataset.

\subsection{Student model}\label{subsec:student}
The LiSER student model accepts log Mel-spectrograms of speech waveforms as inputs. Its architecture is inspired by the 2D CNN LSTM network proposed in~\cite{Zhao_2019}. LiSER's student model consists of three two-dimensional convolution blocks, each with 64 filters, to capture local spatio-temporal features from the spectrograms, followed by an LSTM layer to capture the global context. Additionally, the model includes three multi-layer perceptrons (MLPs) to facilitate the training of the student using various loss functions, which will be detailed in the next subsection. The exact configuration of each component of the model is outlined in Table~\ref{tab:student_model}. In the table, $K_S$ represents the number of emotion categories seen in the available labeled speech, while $K_V$ indicates the number of emotion categories in the DFER dataset which was utilized to train the video teacher.

\begin{table}[h]
\caption{Details of the LiSER student architecture.}
\label{tab:student_model}
\centering
\begin{tabular}{c c c }
\hline
\textbf{Block}        & \textbf{Type}                              & \textbf{Configuration} \\
\hline
\hline
\multirow{3}{*}{1-3}  & Conv2D & Kernel:(3,3) MaxPool:(2,2)\\ \cline{3-3} 
                      & + BatchNorm                                           & Kernel:(3,3) MaxPool:(4,2)\\ \cline{3-3} 
                      & + ReLU                                            & Kernel:(3,1) MaxPool:(4,1)\\ \hline
4                     & LSTM                                       & Hidden size: 64 \\ \hline
5                     & supervised-mlp                             & layers: 1, nodes: $K_S$ \\ \hline
6                     & speech-distill-mlp                         & layers: 2, nodes: 32, $K_S$ \\ \hline
7                     & video-distill-mlp                          & layers: 2, nodes: 32, $K_V$ \\ \hline\hline
\multicolumn{3}{c}{\# Parameters: 105K} \\ \hline
\end{tabular}
\end{table}
\vspace{-0.5mm}
\subsection{Training framework}\label{subsec:train_framework}
Let $\mathcal{D}^L = \{(s_i^L, y_i^L)\}$ denote the labeled speech emotion dataset, where $s_i^L$ represents speech samples and $y_i^L$ represents their emotion labels. $\mathcal{D}^U = \{(s_j^U, v_j^U)\}$ denotes the unlabeled audio-visual dataset, consisting of speech samples $s_j^U$ and their corresponding face image sequences $v_j^U$. We train the student model with parameters $\theta$, by employing standard supervised learning on $\mathcal{D}^L$ and softmax-level distillation-based learning using $\mathcal{D}^U$. The student model consists of three distinct MLP heads namely, $g^{\text{sup}}, g^{\text{sd}}$ and $g^{\text{vd}}$, for the tasks of supervised learning, speech distillation, and video distillation, respectively. Let $t_s$ and $t_v$ represent the networks of speech and video teachers. The loss terms associated with the three tasks are defined as:
\begin{align}
    \mathcal{L}^{\text{sup}}(s_i^L, y_i^L) &= \mathcal{L}^{\text{CE}}\big(g^{\text{sup}}(s_i^L, \theta), y_i^L\big) \label{equation:L_sup} \\
    \mathcal{L}^{\text{sd}}(s_j^U) &= \mathcal{L}^{\text{MAE}}\big(g^{\text{sd}}(s_j^U, \theta), t_s(s_j^U)\big) \label{equation:L_sd}\\
    \mathcal{L}^{\text{vd}}(s_j^U, v_j^U) &= \mathcal{L}^{\text{MAE}}\big(g^{\text{vd}}(s_j^U, \theta), t_v(v_j^U)\big) \label{equation:L_vd}
\end{align}
$\mathcal{L}^{\text{CE}}$ in equation~\eqref{equation:L_sup} represents the cross-entropy loss and $\mathcal{L}^{\text{MAE}}$ in equation~\eqref{equation:L_sd} and~\eqref{equation:L_vd} refers to the mean absolute error (MAE) between the softmax outputs of the student and teacher models. The parameters $\theta$ are learned by minimizing the mini-batch loss defined in the following subsections. Finally, after the model is trained, we utilize $g^{\text{sup}}$ MLP head to make emotion predictions for any given speech signal.

\paragraph*{Mini-batch loss}
Let $\mathcal{L}^{\text{sup}}_i$ represent the supervised loss term for the $i^{\text{th}}$ data point from the labeled dataset $\mathcal{D}^L$ and $\mathcal{L}^{\text{sd}}_j, \mathcal{L}^{\text{vd}}_j$ denote the distillation loss terms for the $j^{\text{th}}$ data point from the unlabeled dataset $\mathcal{D}^U$. The overall loss for a mini-batch containing $N_l$ labeled data points and $N_u$ unlabeled data points is defined as follows:
\begin{align}
\mathcal{L}^{\text{batch}} = \frac{\sum_{i=1}^{N_l} \mathcal{L}^{\text{sup}}_i + 
\sum_{j=1}^{N_u} \big( \lambda^{\text{sd}} \cdot \mathcal{L}^{\text{sd}}_j + 
\lambda^{\text{vd}} \cdot \mathcal{L}^{\text{vd}}_j\big)}{N_l + N_u}
\label{equation:batch_loss}
\end{align}
where $\lambda^{\text{sd}}, \lambda^{\text{vd}}$ are the hyperparameters denoting the weights for the sound and visual distillation loss terms.

\paragraph*{Confidence-enhanced mini-batch loss}
In the mini-batch loss defined in \eqref{equation:batch_loss}, we utilize constant weights (i.e., $\lambda^{\text{sd}}, \lambda^{\text{vd}}$) across all unlabeled data points. This approach leads to the student model emphasizing both modalities uniformly across the entire dataset. However, since each data point may contain varying amounts of emotional information in the two modalities, we enhance the mini-batch loss computation by incorporating the confidence of emotion predictions from the teacher models. Specifically, we introduce instance-level weights denoted as $w_j^{\text{sd}}, w_j^{\text{vd}}$. 
\begin{align}
\mathcal{L}^{\text{batch}}_{\text{conf}} &= \frac{\sum_{i=1}^{N_l} \mathcal{L}^{\text{sup}}_i + 
\sum_{j=1}^{N_u} \big( \lambda^{\text{sd}} \cdot w_j^{\text{sd}} \cdot \mathcal{L}^{\text{sd}}_j + 
\lambda^{\text{vd}} \cdot w_j^{\text{vd}} \cdot \mathcal{L}^{\text{vd}}_j\big)}{N_l + N_u}
\label{equation:batch_loss_conf}
\end{align}
The instance-level confidence weights are computed as the maximum probability values associated with the softmax outputs of the respective teacher models for each data point.

\section{Experiments}
\subsection{Datasets}
This work utilizes audio-visual data from MSP-Face~\cite{Vidal_2020} corpus and speech emotion data from SER benchmark datasets namely, RAVDESS~\cite{livingstone2018ryerson} and CREMA-D~\cite{Cao_2014}.

\paragraph*{MSP-Face} is an audio-visual dataset with recordings collected in-the-wild from video-sharing websites. Each recording features an individual facing the camera and discussing various topics from their daily life in a natural and spontaneous manner. The data was gathered from a diverse group of individuals, conveying a wide range of emotions. The dataset includes YouTube links to these videos, although some of them are no longer available. We successfully downloaded 46.55 hours of data from 386 speakers, with 55\% of them being male. Each video has a frame rate of 30 fps, with an average duration of 9.25 seconds. While some videos included emotion annotations, we do not utilize those annotations and treat all available data as unlabeled. 

We extracted and stored the facial regions from each frame of the recordings using the DeepFace toolkit~\cite{serengil2020lightface} for face detection, alignment, and extraction. To reduce the computational load when training the student model, we pre-computed the softmax outputs of the DFER model for all the videos. The video frames are fed to the DFER model using a sliding window with a length and stride of 16.

\paragraph*{RAVDESS} dataset comprises 1,440 audio-visual recordings from 24 professional actors, of whom 12 are male. The actors vocalize two sentences across eight different emotions including neutral, calm, happy, sad, angry, fearful, surprise, and disgust. For our study, we utilize only the speech portion of this dataset to train and evaluate our student model.

\paragraph*{CREMA-D} dataset consists of 7,442 audio-visual clips from a diverse group of 91 actors with 48 of them being male. Each actor spoke from a selection of 12 sentences multiple times, conveying emotions from six categories: anger, disgust, fear, happy, neutral, and sad. As with RAVDESS, we focus only on the speech portion of this dataset in the current study.

\subsection{Development of teacher models}
We developed the speech teacher model by fine-tuning the pre-trained HuBERT for the SER task, utilizing labeled speech samples from the same dataset used to train the student model. The fine-tuning of HuBERT is achieved by applying Low-Rank Adaptation (LoRA)~\cite{hu2022lora} to the weight matrices of the self-attention modules. We utilized 80\% of the labeled speech to fine-tune it for a maximum of 50 epochs and chose the checkpoint corresponding to the epoch with the best SER performance on the remaining 20\% data. This selected checkpoint serves as the speech teacher.

Our video teacher is an S2D model trained in \cite{chen2024static}, using video samples from the FERV39k corpus~\cite{Wang_2022}. The FERV39k dataset comprises videos with a frame rate of 30 fps, spanning seven emotion categories: angry, disgust, fear, happy, neutral, sad, surprise. The S2D model was trained to predict emotions based on any randomly selected 16 consecutive face image frames (equivalent to 0.5s) extracted from these video samples.

\subsection{Input to the student model}\label{subsec:input_student}
The student model receives log Mel-spectrograms derived from speech signals of 3s in duration as its inputs. The Mel-spectrogram is calculated using 64 Mel bands, with a window size of 128 ms and a stride of 32ms. For speech signals shorter than 3s, zero padding is applied before inputting them into the model. For signals exceeding 3s, a random 3-second segment is selected from the entire signal during training and fed into the model. In the evaluation phase, multiple 3s segments are extracted from the entire signal using a sliding window of 3s with a stride of 0.1s. Note that a single prediction is generated for the entire signal by averaging the $g^{\text{sup}}$ logits (ref.~section~\ref{subsec:train_framework}) corresponding to these smaller segments.

\subsection{Mini-batch loss computation}
For labeled data points in the mini-batch, LiSER computes the cross-entropy loss between the emotion labels and the logits from $g^{\text{sup}}$ MLP. For unlabeled data points, the loss terms are computed for both speech and video distillation. As outlined in section~\ref{subsec:input_student}, we feed a 3s speech signal from the unlabeled data point to the student model, obtaining outputs from the relevant MLP heads for the distillation tasks. We then obtain the softmax outputs from the two teacher models by feeding the respective 3s audio and video inputs into them. Since the S2D model can only predict from 0.5s-duration video clips, we calculate the softmax prediction for the entire 3s video by averaging the outputs from all corresponding 0.5s clips. In contrast, HuBERT can handle speech signals of any length. However, to ensure a fair comparison between the knowledge distillation from both modalities, we similarly average the outputs from the 0.5s segments of the 3s speech signal. After computing the relevant loss terms for each data point, we utilize the mini-batch loss defined in equations~\ref{equation:batch_loss} and ~\ref{equation:batch_loss_conf} to train the student model.

\subsection{Training configurations}
The student models are trained under various configurations to assess the effectiveness of different components within our training framework. The performance of these models is presented in Table~\ref{tab:rav_cremad_results}. 
\empfont{no-dstl} indicates the training of the student using only labeled speech data. \empfont{vid-dstl} and \empfont{sp-dstl} refer to the training with supervised learning over labeled speech in conjunction with distillation from either video or speech teacher, respectively. \empfont{vid-sp-dstl} and \empfont{conf-vid-sp-dstl} refers to the training using mini-batch loss specified in equations~\eqref{equation:batch_loss} and~\eqref{equation:batch_loss_conf}, respectively, with $\lambda^{\text{vd}}\neq 0,~\lambda^{\text{sd}} \neq 0$.

\subsection{Experimental results}
\begin{table}[t]
\caption{Speech emotion recognition performance of LiSER student models on RAVDESS and CREMA-D.}
\label{tab:rav_cremad_results}
\centering
\begin{tabular}{c c c c c}
    \hline
     & \multicolumn{2}{c}{\textbf{RAVDESS}} & \multicolumn{2}{c}{\textbf{CREMA-D}} \\ \hline
    \textbf{Configuration} & \textbf{UAR} & \textbf{WAR} & \textbf{UAR} & \textbf{WAR}\\ \hline \hline
    \empfont{no-dstl} & 0.517 & 0.535 & 0.551 & 0.55 \\ 
    \empfont{vid-dstl} & 0.534 & 0.547 & 0.576 & 0.575 \\ 
    \empfont{sp-dstl} & 0.545 & 0.556 & 0.576 & 0.575 \\ 
    \empfont{vid-sp-dstl} & 0.556 & 0.57 & \textbf{0.584} & \textbf{0.583} \\ 
    \empfont{conf-vid-sp-dstl} & \textbf{0.595} & \textbf{0.611} & 0.578 & 0.576 \\ 
    \hline
\end{tabular}
\end{table}
We evaluate the LiSER framework on RAVDESS and CREMA-D using Unweighted Average Recall (UAR) and Weighted Average Recall (WAR). We follow a five-fold cross-validation protocol to divide the labeled dataset into train, validation and test sets, ensuring no overlap in speakers across these sets. The resulting training set is augmented with samples from MSP-Face when training with distillation loss terms. In each training configuration, we train the student model for a maximum of 50 epochs, selecting the checkpoint corresponding to the epoch with best validation set performance. The validation set is used to determine the optimal values for $\lambda^{\text{vd}},~\lambda^{\text{sd}}$ over $\{0.1, 0.5, 1, 5, 10\}$. The student models are trained using AdamW optimizer with a learning rate of 1e-4, batch size of 25. 

The results in Table~\ref{tab:rav_cremad_results} show that using knowledge from speech and video teachers enhances the performance of the student model. In RAVDESS, when comparing with the \empfont{no-dstl} scenario, we see improvements of 3.29\% and 5.42\% in UAR from the video and speech teachers, respectively. For CREMA-D, both teachers lead to a 4.54\% improvement. Combining both speech and video distillation in the \empfont{vid-sp-dstl} approach gives even better results, with increases of 7.54\% in RAVDESS and 5.99\% in CREMA-D. We also looked at how incorporating teacher models’ confidence in emotion predictions affects results. In RAVDESS, this integration improves UAR by 15.09\% compared to the \empfont{no-dstl} approach. However, in CREMA-D, the improvement slightly decreases from 5.99\% to 4.9\%.

We also conduct an ablation study to examine the effects of different loss functions and training methodologies for knowledge transfer from the teacher models, as shown in Table~\ref{tab:rav_results}. In distill-ce, we replaced the MAE distillation loss with cross-entropy (CE) loss. We compared LiSER's training method, which uses both labeled and unlabeled data at the same time, with the two-stage training method used in~\cite{goncalves2022improving,Shukla_2023,li2021speech}. The two-stage method first trains on unlabeled data, followed by fine-tuning with labeled data. Our results show that MAE outperforms CE loss, which aligns with related research such as~\cite{Ghosh_2017} which finds MAE more resilient to noisy labels. Additionally, LiSER's training method outperforms the two-stage training. 

\begin{figure}[t]
  \centering
  \includegraphics[width=0.35\textwidth]{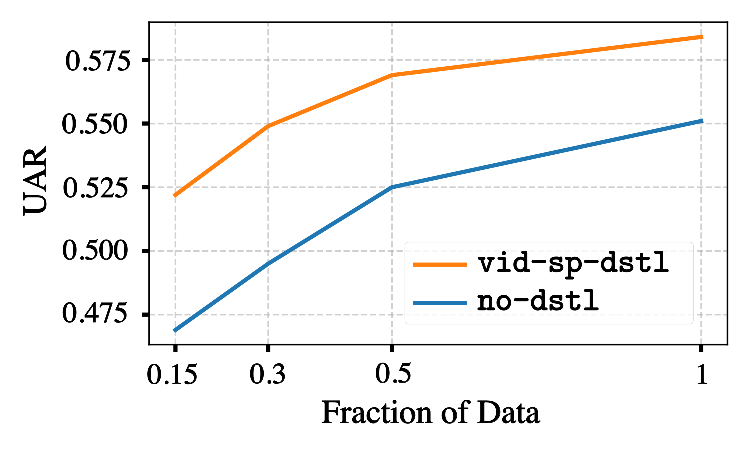}
  \caption{Ablation study over CREMA-D on the impact of using less labeled data in the training phase.}
  \label{fig:small_cremad_results}
\end{figure}

Finally, we assessed the impact of using less labeled speech data in the \empfont{vid-sp-dstl} scenario by training the student model on smaller subsets of the labeled data. Figure~\ref{fig:small_cremad_results} displays these results for CREMA-D. The findings indicate that the student model trained with all data from MSP-Face and only half of the labeled data performs better than the model that used the whole labeled dataset in the \empfont{no-dstl} scenario.

\begin{table}[t]
\caption{Ablation study of distillation loss and training methodology on RAVDESS.}
\label{tab:rav_results}
\centering
\begin{tabular}{c c c c c}
    \hline
     & \multicolumn{2}{c}{\textbf{Video}} & \multicolumn{2}{c}{\textbf{Speech}} \\
    \hline
    \textbf{Configuration} & \textbf{UAR} & \textbf{WAR} & \textbf{UAR} & \textbf{WAR}\\ \hline \hline
    \empfont{vid-dstl} / \empfont{sp-dstl} & \textbf{0.534} & \textbf{0.547} & \textbf{0.545} & \textbf{0.556}\\
    distill-ce & 0.523 & 0.54 & 0.54 & 0.546\\
    two-stage-train & 0.464 & 0.485 & 0.511 & 0.528
    \\\hline
\end{tabular}
\end{table}

\section{Conclusion}
This paper developed a knowledge distillation framework called LiSER that improves lightweight models for recognizing emotions in speech by using unlabeled audio-visual data. We validated this framework with an unlabeled audio-visual dataset collected in-the-wild. Our results show significant improvements of up to 15.09\% and 5.99\% in unweighted average recall on RAVDESS and CREMA-D benchmarks, respectively. The findings indicate that the knowledge gained from teacher models which understand speech emotions and facial expressions, enhances the performance of the student models. Moreover, simultaneous distillation from both audio and visual modalities yields better results than using a single modality. The results from RAVDESS also suggest that integrating confidence measures from teachers' predictions can help each data point to effectively utilize the varying levels of information offered by different teacher models.

\bibliographystyle{IEEEtran}
\bibliography{template}

\end{document}